%% file: main.tex
\documentclass[10pt,twocolumn,letterpaper]{article}

\usepackage{cvpr}              

\input{preamble}

\definecolor{cvprblue}{rgb}{0.21,0.49,0.74}
\usepackage[pagebackref,breaklinks,colorlinks,citecolor=cvprblue]{hyperref}
\usepackage{fontawesome5}

\title{ZARRIO @ Ego4D Short Term Object Interaction Anticipation Challenge: Leveraging Affordances and Attention-based models for STA}

\author{Lorenzo Mur-Labadia*\\
University of Zaragoza\\
{\tt\small lmur@unizar.es}
\and
Ruben Martinez-Cantin\\
University of Zaragoza\\
{\tt\small rmcantin@unizar.es}
\and
Josechu Guerrero\\
University of Zaragoza\\
{\tt\small jguerrer@unizar.es}
\and
Giovanni Maria Farinella\\
University of Catania\\
{\tt\small giovanni.farinella@unict.it}
}

\begin{document}
\maketitle
\input{sec/0_abstract}

\input{sec/1_intro}
\input{sec/2_formatting}

\input{sec/3_finalcopy}
{
    \small
    \bibliographystyle{ieeenat_fullname}
    \bibliography{main}
}


\end{document}

%% file: preamble.tex
\usepackage[dvipsnames]{xcolor}


%% file: sec/0_abstract.tex
\begin{abstract}

Short-Term object-interaction Anticipation (STA) consists of detecting the location of the next-active objects, the noun and verb categories of the interaction, and the time to contact from the observation of egocentric video.
We propose STAformer, a novel attention-based architecture integrating frame-guided temporal pooling, dual image-video attention, and multi-scale feature fusion to support STA predictions from an image-input video pair.
Moreover, we introduce two novel modules to ground STA predictions on human behavior by modeling affordances. First, we integrate an environment affordance model which acts as a persistent memory of interactions that can take place in a given physical scene. Second, we predict interaction hotspots from the observation of hands and object trajectories, increasing confidence in STA predictions localized around the hotspot. On the test set, our results obtain a final 33.5 N mAP, 17.25 N+V mAP, 11.77 N+$\delta$ mAP and 6.75 Overall top-5 mAP metric when trained on the v2 training dataset. \href{https://github.com/lmur98/AFFttention}{We will release the code and pre-extracted affordances. \cite{affttention}} 
\end{abstract}

%% file: sec/1_intro.tex
\section{Introduction}
\label{sec:intro}

Predictive capabilities are essential for assistive egocentric devices. For instance, an intelligent wearable device could preemptively warn an electrician about potential switchboard short-circuits. The proposed Short-Term Object Interaction Anticipation (STA) task consists in, given an input video, anticipate the next action and active object category, the object's bounding box, an the time to contact. Inspired by this challenge, the community proposed different approaches~\cite{chen2022internvideo, tong2022videomae, pasca2023summarize, ragusa2023stillfast, thakur2023enhancing, thakur2023guided, thakur2024leveraging}.

This technical report is a shortened version of a more complete work \cite{affttention}.
Our aim with this work is to advance research in STA with two main contributions. 
First, we propose STAformer, a principled architecture unifying the computation of image and video inputs with attention-based components. Differently from previous approaches~\cite{grauman2022ego4d,thakur2023guided,pasca2023summarize}, we explicitly designed STAformer to operate on an image-video input pair, which is specific to the considered STA task. Our architecture is a significant departure from convolutional baselines~\cite{grauman2022ego4d,ragusa2023stillfast} and aims to offer the convenience and state-of-the-art performance of attention-based feature extractors~\cite{oquab2023dinov2,bertasius2021space} and components.
Next, we propose two modules to ground predictions into human behavior by modeling affordances.
We first leverage environment affordances, estimated by matching the input observation to a learned affordance database, to predict probability distributions over nouns and verbs, which are used to refine verb and noun probabilities predicted by STAformer. Our intuition is that linking a zone across similar environments captures a description of the feasible interactions, grounding predictions into previously observed human behavior.
The second affordance module aims to relate STA predictions to a spatial prior of where an interaction may take place in the current frame. This is done by predicting an interaction hotspot~\cite{liu2022joint}, which is used to re-weigh confidence scores of STA predictions depending on the object's location.

Experiments on Ego4D~\cite{grauman2022ego4d} obtain on the test set a final 33.5 N mAP, 17.25 N+V mAP, 11.77 N+$\delta$ mAP and 6.75 Overall top-5 mAP metric when trained on the v2 training dataset. On the validation split, we obtain a relative gain of +43.4 $\%$ N mAP, + 47.6 $\%$ N+V mAP, +36.5 $\%$ N+$\delta$ mAP and + 42.1 $\%$ Overall top-5 metric when compared with the last year's winner of the STA challenge \cite{thakur2023guided}. We also detailed the contribution of each component through ablations and showed that the integration of affordances is beneficial independently of the model performance.

%% file: sec/2_formatting.tex
\begin{figure*}[t]
\centering
\includegraphics[width=\textwidth]{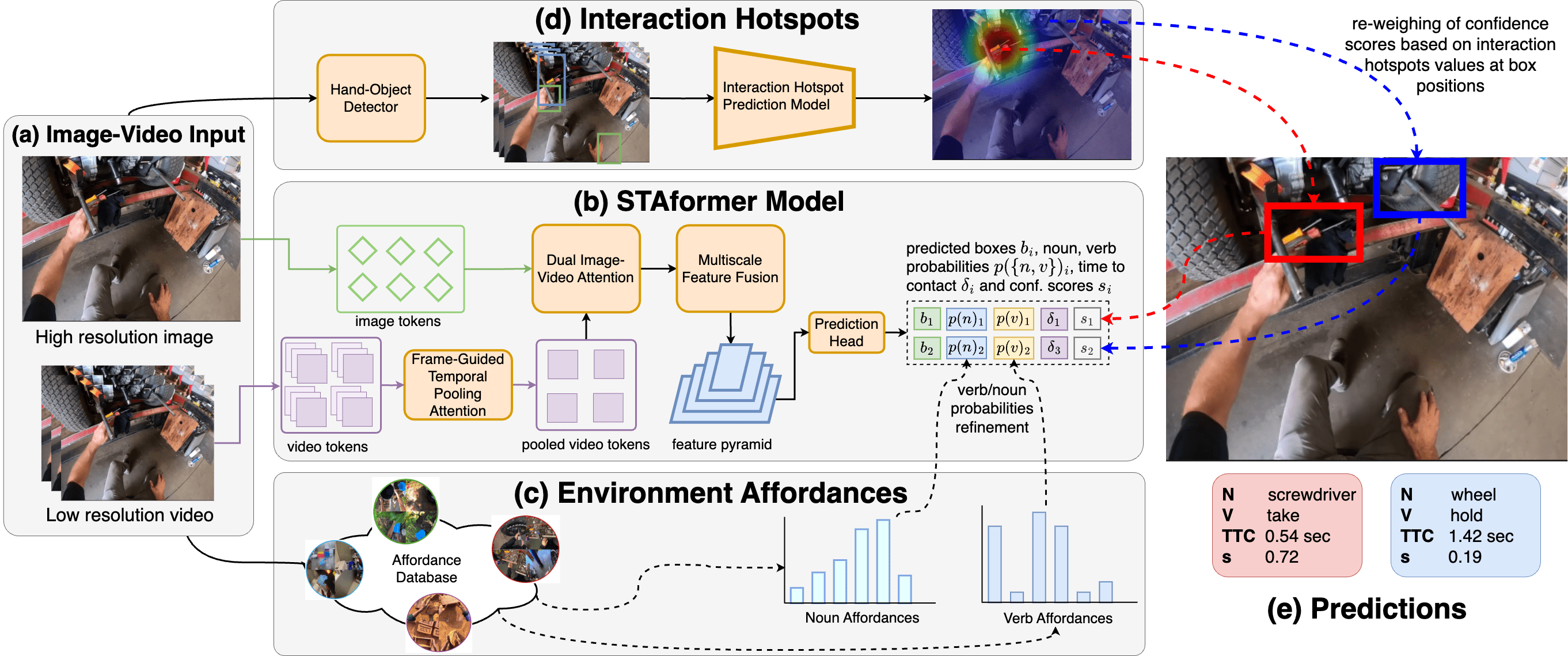}
\caption{(a-b) The image-video pair input is processed by the proposed STAformer model which predicts object bounding boxes, the associated verb/noun probabilities, time-to-contact and confidence scores.(c) Environment affordances are inferred from video and used to refine the predicted noun/verb probabilities. (d) Our model observes detected hand-object interactions in the video and predicts an interaction hotspot probability map, which is used to re-weigh confidence scores based on box locations, leading to (e) our final predictions.} 
\label{fig:encoder}
\end{figure*}

\section{Methods}
\label{sec:methods}

\subsection{STAformer Architecture}

\textbf{Feature Extraction:}
We follow previous work~\cite{grauman2022ego4d,ragusa2023stillfast} and process a high resolution image $I_T \in \mathbb{R}^{h_s \times w_s \times 3}$ sampled from the input video $\mathcal{V}_{:T}$ at time $T$ and a sequence of low-resolution frames $\mathcal{V}_{T - t:T} \in \mathbb{R}^{t \times h_f \times w_f \times 3}$ taken $t$ time-steps before time $T$. First, we extract high-resolution 2D features from the $I_T$ input with a DINOv2 model\cite{oquab2023dinov2}, obtaining a set of 2D image tokens $\Phi_{2D}(I_T)$ and a class token $C_I$.
We also extract spatio-temporal 3D features from the $\mathcal{V}_{T - t: T}$ input with a TimeSformer model~\cite{bertasius2021space} in the form of video tokens $\Phi_{3D}(\mathcal{V}_T)$ and a class token $C_\mathcal{V}$.

\textbf{Frame-guided Temporal Pooling Attention:}
While the overall video tokens provide a spatio-temporal representation of the input video, STA predictions need to be aligned to the spatial location of the last video frame. The frame-guided temporal pooling attention maps video tokens to the spatial reference system of the last video frame, compressing the 3D representation obtained by the TimeSformer to a 2D one. The 3D video tokens $\Phi_{3D}(\mathcal{V}_{T-t:T})$ are mapped to 2D pooled video tokens denoted as $\Phi_{3D}^{2D}(\mathcal{V}_{T-t:T})$ adopting a residual multi-head cross-attention mechanism. 
Used as queries, last-frame tokens guide an adaptive temporal pooling that summarizes the spatio-temporal feature map computed by the TimeSformer model and maps it to the 2D reference space of the last observed frame. 


\textbf{Dual Image-Video Attention fusion}
Image tokens $\Phi_{2D}(I_T)$ and pooled video tokens $\Phi_{3D}^{2D}(\mathcal{V}_{T-t:T})$ are spatially aligned, but carry different information, with image tokens encoding fine-grained visual features and video tokens encoding scene dynamics, so we adopt a dual cross-attention module for refining both modalities.
This module adopts a residual dual cross-attention that aims to enrich image tokens with scene dynamics information coming from video tokens through image-guided cross-attention.
The video-guided cross-attention works in a similar way to compute refined video tokens $\tilde{\Phi}_{3D}(\mathcal{V}_{T-t:T})$, but queries are computed from video tokens while keys and values are computed from image tokens.

\textbf{Feature Fusion and prediction head:}
Refined image and video class tokens are summed to obtain the overall class token $C_T = \tilde{C}_{I} + \tilde{C}_{\mathcal{V}}$.
Refined image $\tilde{\Phi}_{2D}(I_T)$ and video $\tilde{\Phi}_{3D}(\mathcal{V}_{T-t:T})$ tokens are mapped to respective multi-scale feature pyramids~\cite{lin2017feature} using bilinear interpolation. The two feature pyramids are summed and passed through a 2D $3 \times 3$ convolution to obtain the fused feature pyramid. We adopt the prediction head proposed in~\cite{ragusa2023stillfast} to obtain the final predictions $(\hat b_i, \hat n_i, \hat v_i, \hat \delta_i, \hat s_i)$.

\begin{table*}[]
    \centering
    
    \captionsetup{font=normal}
    \caption{Ablation study of the different components of STAformer on the v1-val split of Ego4D. \faSnowflake 
 Encoder frozen \faWrench  Encoder last-blocks fine-tuned \faCog  Full encoder trained.}
    \label{tab:mAP_v1} 
    \begin{tabular}{|c|cccc|cccc|}

    \hline
    Exp. & Image Encoder & Video Encoder & Temporal pooling & 2D-3D Fusion & N & N + V & N + $\delta$ & All \\ \hline
    \cite{ragusa2023stillfast}&R50 \faCog  & X3D \faCog  & Mean & Sum & 16.21 & 7.52 & 4.94 & 2.48 \\ \hline 
         A & DINOv2  \faSnowflake & - & - & - & 17.48 & 8.64 & 5.20 & 2.52 \\
         B & DINOv2 \faSnowflake & TimeSformer \faWrench & Mean & Sum & 16.67 & 8.38 & 5.16 & 2.63 \\
         C & DINOv2  \faSnowflake & TimeSformer \faWrench & Frame-guided & Sum & 19.78 & 10.04 & 6.35 & 3.39 \\ 
         D & DINOv2  \faWrench & TimeSformer \faWrench & Frame-guided & Dual $I \leftrightarrow \mathcal{V}$ & 21.71 & 10.75 & 7.24 & 3.53 \\
         E & DINOv2  \faWrench & TimeSformer \faWrench & MH-Frame-guided & MH-Dual $I \leftrightarrow \mathcal{V}$ & \textbf{23.02} & \textbf{11.57} & \textbf{7.86} & \textbf{3.85} \\ \hline
    \end{tabular}%
\label{ref:mAP_v1}
\end{table*}

\subsection{Leveraging environment affordances.}

We build a robust representation of the environment affordances \cite{nagarajan2020ego} from the observation of human activities in video. It encapsulates a powerful inductive bias to predict
the interaction that the user is going to perform next. We first build an affordance database grouping the training videos according to their visual similarity in activity-centric zones. At inference time, we match a novel video $\mathcal{V}'$ to the most functionally similar zones in the affordance database, estimating the distribution of the affordable interactions in the new video. The affordance probability distribution is obtained by weighting the counts of nouns/verbs present in the top-K nearest zones according to the respective visual and narrative similarity. We use the nouns and verbs affordance distributions to refine the respective nouns and verbs probabilities predicted by STAformer.

\subsection{Leveraging interaction hotspots.}
While our affordance database gives us information on which objects (nouns) and interaction modes (verbs) are likely to appear in the current scene, it does not give us any information on \textit{where} the interaction will take place in the observed images.
As noted in previous works~\cite{liu2022joint}, observing how hands move in egocentric videos can allow us to predict the interaction hotspot~\cite{liu2022joint,nagarajan2020ego}.
We exploit this concept and include a module to predict an interaction hotspot by observing frames, hands, and objects. We hence re-weigh the confidence scores $s_i$ of STA predictions according to the location of the respective bounding box centers in the predicted interaction hotspot, to reduce the influence of false positive detections falling in areas of unlikely interaction.

%% file: sec/3_finalcopy.tex
\section{Results}

\begin{table}[t]
    \centering
        \captionsetup{font=normal}
            \caption{Results in mAP on the validation split of Ego4D-STA v2.  \textbf{Best results} in bold. Relative gain is with respect to \underline{second best}.}
            \label{tab:mAP_v2}
        \begin{tabular}{|c|cccc|}
            \hline
            Model & N & N + V & N +$\delta$ & All \\ \hline
            FRCNN+SF~\cite{grauman2022ego4d} & 21.00 & 7.45& 7.07 & 2.98 \\
            InternVideo~\cite{chen2022internvideo} & 19.45 & 8.00 & 6.97 & 3.25 \\
            StillFast~\cite{ragusa2023stillfast} & 20.26 & 10.37 & 7.26 & 3.96 \\
            GANO v2~\cite{thakur2023guided} & \underline{20.52} & \underline{10.42} & \underline{7.28} & \underline{3.99} \\ \hline
            STAformer  & 24.85 & 13.45 & 7.41 & 4.90 \\
            STAformer+AFF & 27.03 & 14.36 & 8.72 & 5.04 \\
            STAformer+MH & 27.51 & 14.68 & 9.63 & 5.50 \\
            STAformer+MH+AFF & \textbf{29.39} & \textbf{15.38} & \textbf{9.94} & \textbf{5.67} \\
            \hline
            Gain (rel $\%$) & \textcolor{ForestGreen}{+43.3}& \textcolor{ForestGreen}{+47.6}& \textcolor{ForestGreen}{+36.5}& \textcolor{ForestGreen}{+42.1}\\\hline
        \end{tabular}
       
\end{table}

\begin{table}[t]
    \centering
       \captionsetup{font=normal}
        \caption{Results in mAP on the test split of Ego4D-STA with models trained on v2 split.}
        \label{tab:mAP_test}
        \begin{tabular}{|c|cccc|}
        \hline
        Model  &  N & N + V & N + $\delta$ & All \\ 
         \hline
        StillFast \cite{ragusa2023stillfast}   & 25.06 & 13.29 & 9.14 & 5.12 \\
        GANO v2 \cite{thakur2023guided}  & 25.67 & \underline{13.60} & 9.02 & 5.16 \\
        Language NAO  & \underline{30.43} & 13.45 & \underline{10.38} & \underline{5.18} \\ \hline
        STAformer & 30.61 & 16.67 & 10.06 & 5.62 \\
        STAformer+AFF  & 32.39 & 17.38 & 10.26 & 5.70 \\
        STAformer+MH & 31.99 & 16.79 & 11.62 & 6.72 \\
        STAformer+MH+AFF &  \textbf{33.50} & \textbf{17.25} & \textbf{11.77} & \textbf{6.75} \\\hline
        \end{tabular}
        
\end{table}

We evaluate our approach on the challenging STA-Ego4D benchmark ~\cite{grauman2022ego4d}. We adopt standard Noun (N), Noun+Verb (N+V), Noun+time-to-contact (N+$\delta$) and Noun+Verb+time-to-contact (All) Top-5 mean Average Precision (mAP). Table~\ref{tab:mAP_v1} details an ablation study on the different architectural components of STAformer. While \cite{ragusa2023stillfast} fully trains both image-video encoders, our Experiment A keeps the 2D encoder freeze and trains solely the STA prediction head, reflecting the modeling capacity of DINOv2 features. The naive addition of video features without any temporal module (Experiment B) does not show noticeable improvements. Experiment C measures the influence of the frame-guided temporal pooling and its spatio-temporal video understanding capacity. Using a dual cross-attention fusion between the image and video tokens in Experiment D suggests the importance of the refinement of both modalities. Finally, incorporating multi-head attention (Experiment E) produces a consistent improvement in all the metrics due to its ability to capture diverse patterns from multiple representations simultaneously.

\begin{figure*}
    \centering
    \includegraphics[width=0.19\textwidth]{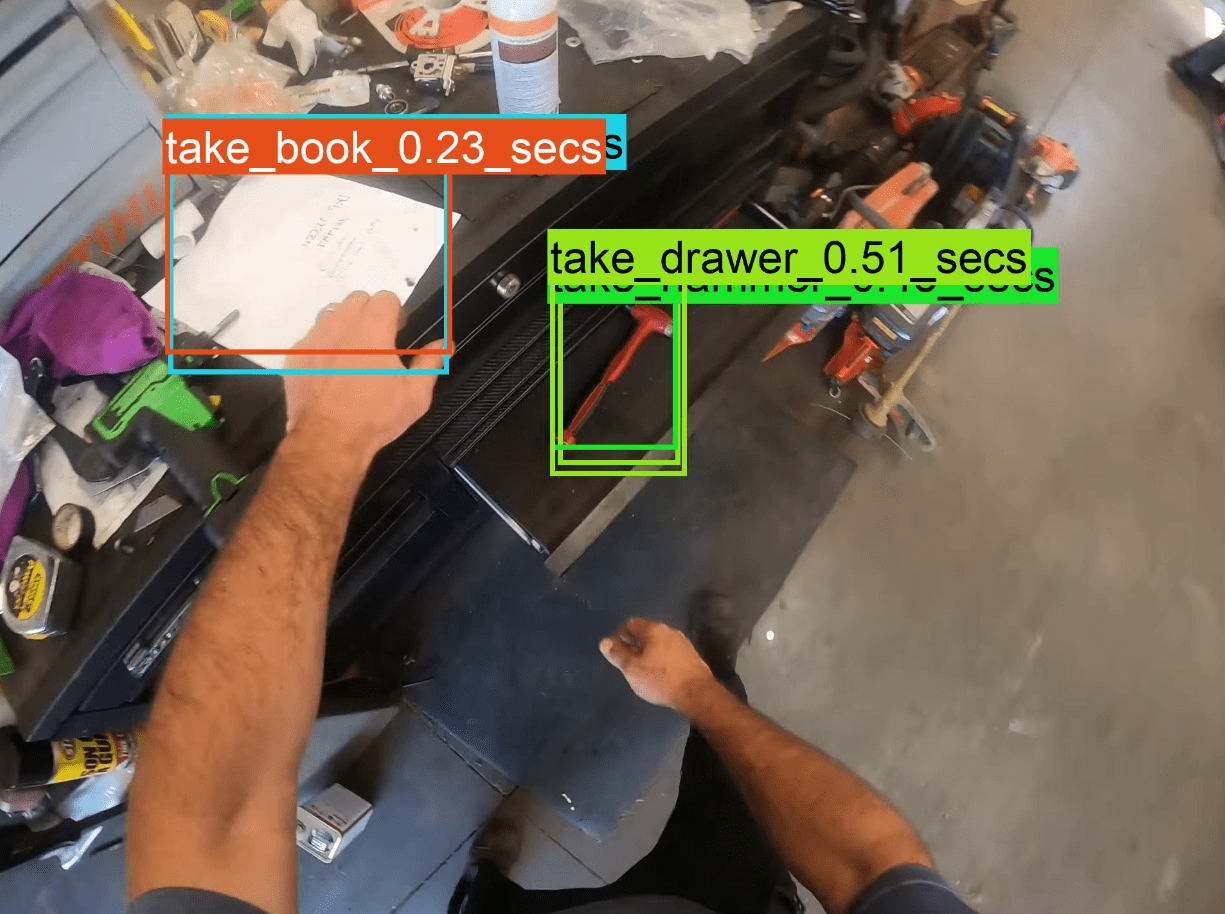}
    \includegraphics[width=0.19\textwidth]{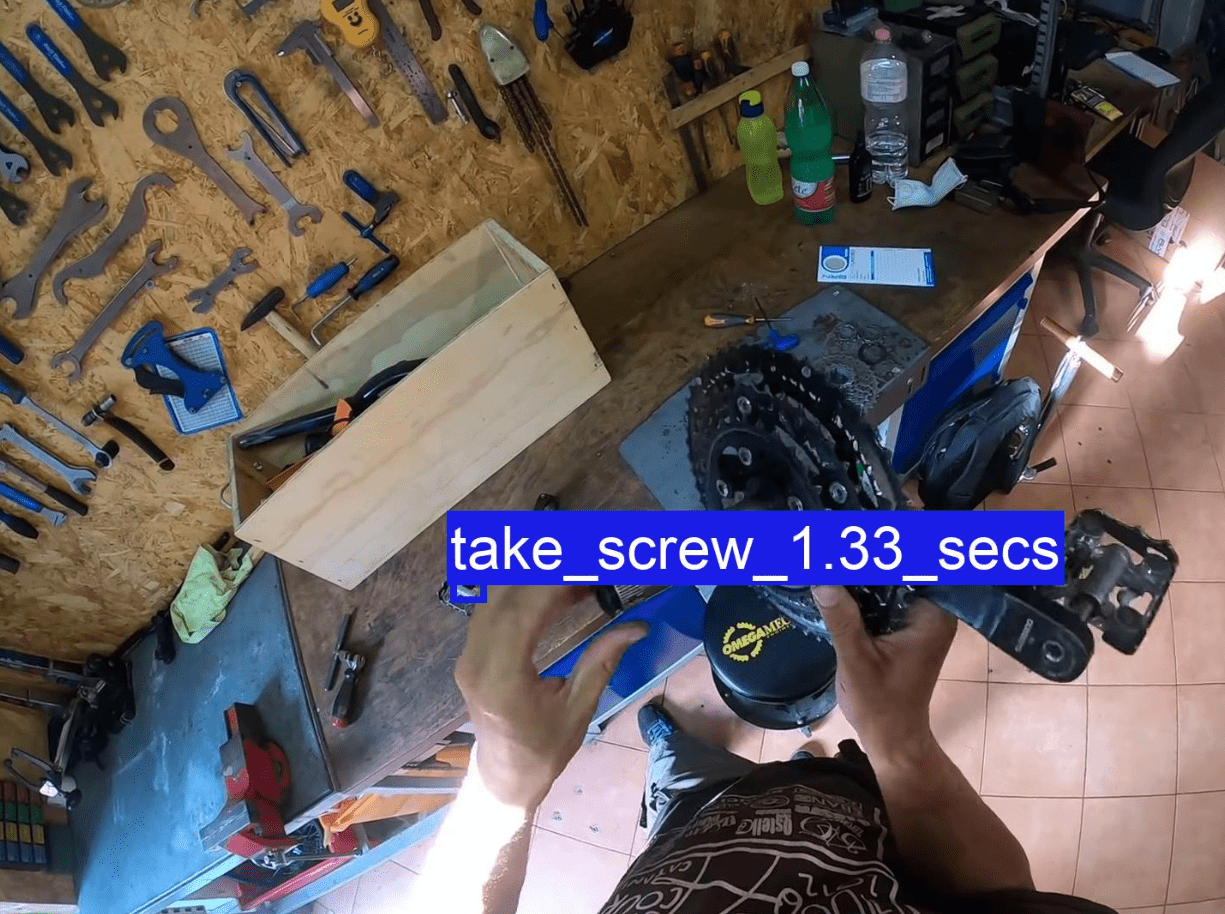}
    \includegraphics[width=0.19\textwidth]{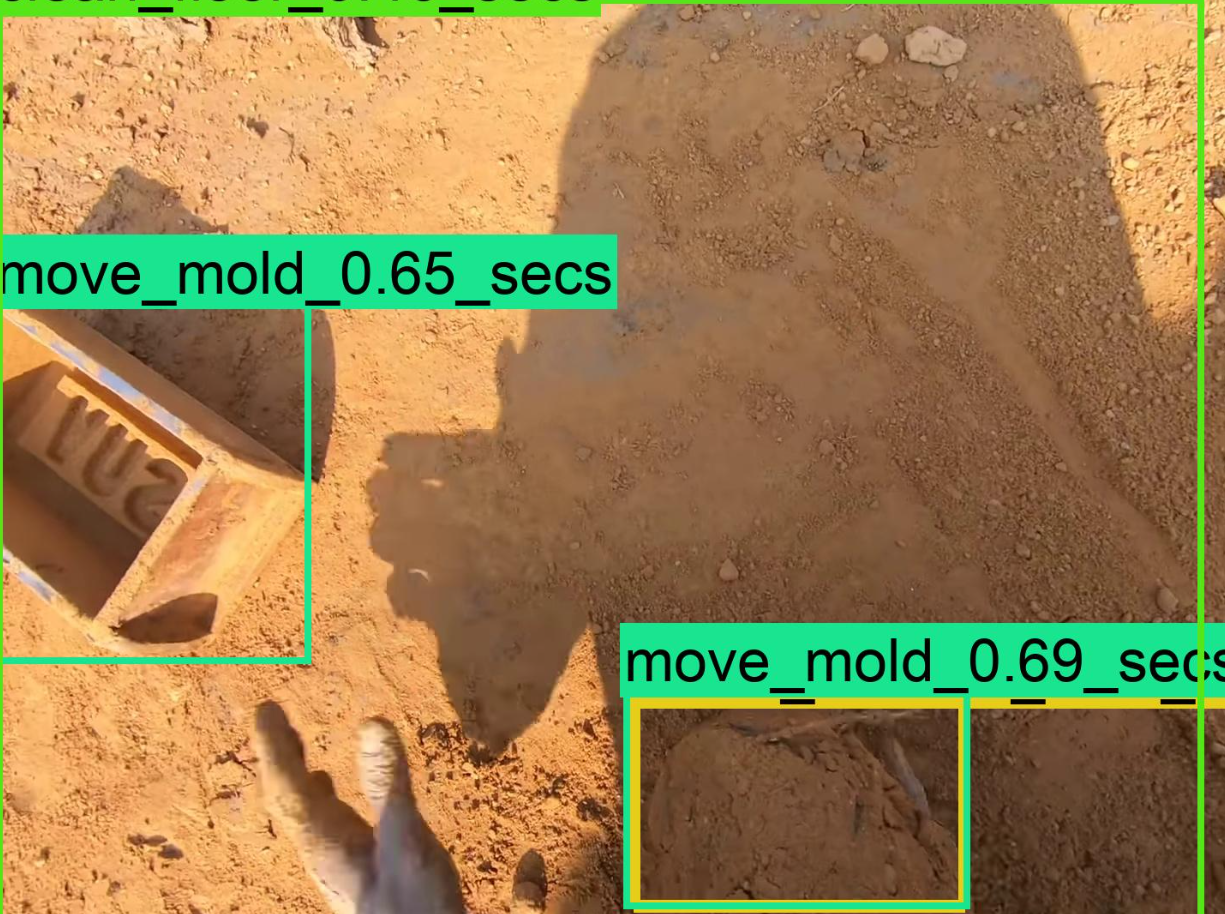}
    \includegraphics[width=0.19\textwidth]{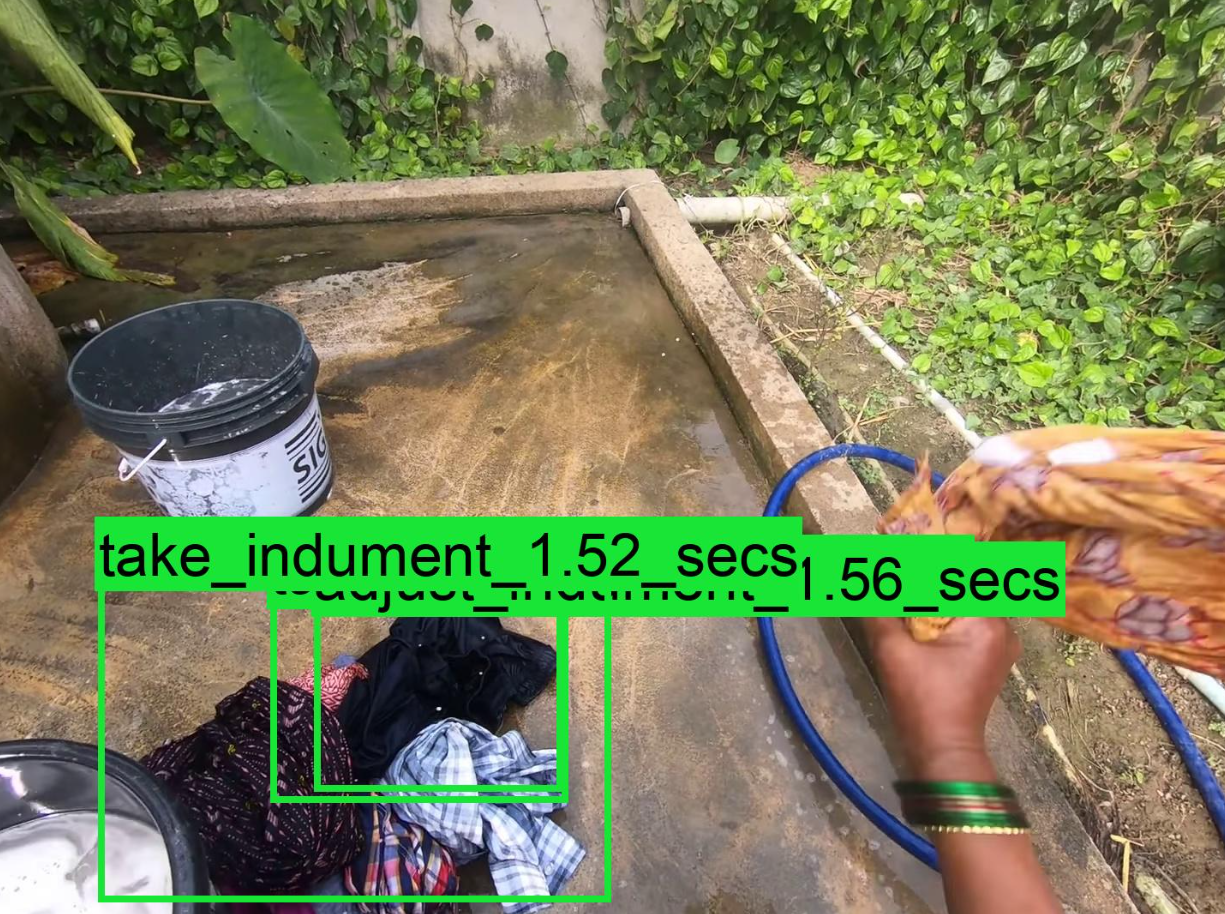}
    \includegraphics[width=0.19\textwidth]{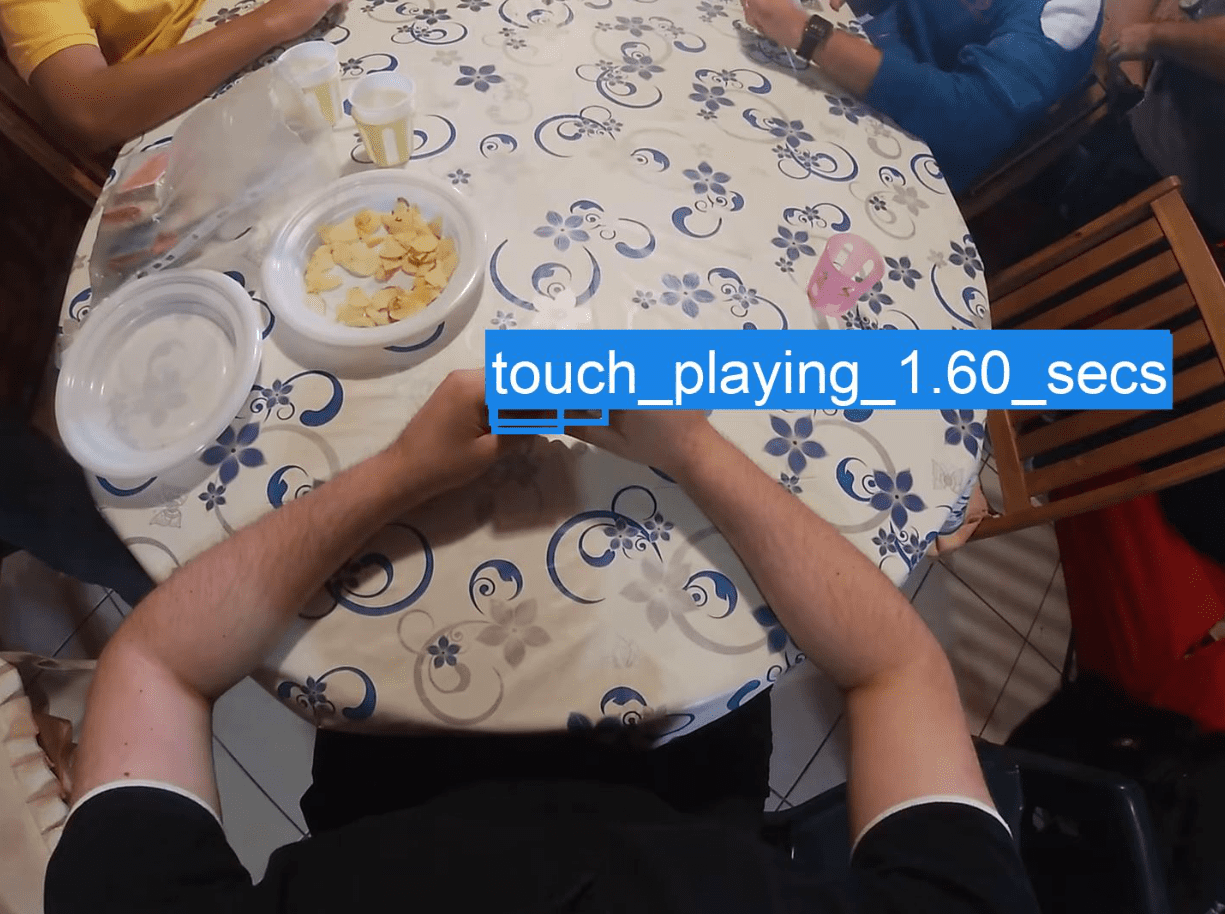}
    \captionsetup{font=normal}
    \caption{Qualitative results on Ego4D.}
    \label{fig:supp_ql_EGO4D}
\end{figure*}

We report the performance of our model on the validation split of Ego4D-STA v2 on Table~\ref{tab:mAP_v2}, showing a significant improvement with last year's winner \cite{thakur2023enhancing}. Following, Table~\ref{tab:mAP_test} reports the results of the test split of Ego4D with methods trained on the v2 split. Our method achieves a 33.5 N mAP, 17.25 N+V mAP, 11.77 N+$\delta$ mAP, and 6.75 Overall top-5 mAP metric. In both cases, we detail an ablation study on the influence of the affordances on different base architectures. In all the cases, the joined semantic generalization capacity of environment affordances and the spatial refinement of the interaction hotspots improve all the metrics, especially the N+V mAP. We finally report some qualitative results in Figure \ref{fig:supp_ql_EGO4D}. The modeling capacity of the high-resolution DINOv2 features helps the model to detect small objects like \textit{screw} or \textit{playing cards}. The model struggles to detect in uncertain situations with multiple objects \textit{take indument} or \textit{move mold}.

\section{Conclusions}

In this technical report, we addressed the problem of Short-Term object-interaction Anticipation (STA). Our main contributions are STAformer, a novel attention-based architecture for STA, and the integration of affordances to ground STA predictions into human behavior. Our results 
obtain a 33.5 N mAP, 17.25 N+V mAP, 11.77 N+$\delta$ mAP, and 6.75 Overall top-5 mAP metric on the test set. 

\section*{Acknowledgements}

Research at University of Catania has been supported by the project Future Artificial Intelligence Research (FAIR) – PNRR MUR Cod. PE0000013 - CUP. E63C22001940006. Research at the University of Zaragoza was supported by the projects PID2021-125209OB-I00 and TED2021-129410B-I00 (MCIN/AEI/10.13039/501100011033 and
NextGenerationEU/PRTR) and the Aragon Government (DGA-T45\textunderscore23R).